**Analyzing Robustness of the Deep Reinforcement Learning Algorithm in Ramp Metering Applications Considering False Data Injection Attack and Defense**

*Corresponding author: Diyi Liu
**Diyi Liu**
Affiliations: Department of Civil and Environment Engineering
University of Tennessee, Knoxville, Tennessee, USA
Email: dliu27@vols.utk.edu

**Lanmin Liu**
Affiliations: Department of Civil and Environment Engineering
University of Tennessee, Knoxville, Tennessee, USA
Email: lliu53@vols.utk.edu

**Lee D Han**
Affiliations: Department of Civil and Environment Engineering
University of Tennessee, Knoxville, Tennessee, USA
Email: lhan@utk.edu

Word Count: 3,635 words + 3 Tables (250 words per table) = 4385

*Submitted [Aug 1st, 2023]*

*Diyi Liu, Lanmin Liu, and Lee D Han*

**ABSTRACT**
Ramp metering is the act of controlling on-going vehicles to the highway mainlines. Decades of practices of ramp metering have proved that ramp metering can decrease total travel time, mitigate shockwaves, decrease rear-end collisions by smoothing the traffic interweaving process, etc. Besides traditional control algorithm like ALINEA, Deep Reinforcement Learning (DRL) algorithms have been introduced to build a finer control. However, two remaining challenges still hinder DRL from being implemented in the real world: (1) some assumptions of algorithms are hard to be matched in the real world; (2) the rich input states may make the model vulnerable to attacks and data noises. To investigate these issues, we propose a Deep Q-Learning algorithm using only loop detectors information as inputs in this study. Then, a set of False Data Injection attacks and random noise attack are designed to investigate the robustness of the model. The major benefit of the model is that it can be applied to almost any ramp metering sites regardless of the road geometries and layouts. Besides outcompeting the ALINEA method, the Deep Q-Learning method also shows a good robustness through training among very different demands and geometries. For example, during the testing case in I-24 near Murfreesboro, TN, the model shows its robustness as it still outperforms ALINEA algorithm under Fast Gradient Sign Method attacks. Unlike many previous studies, the model is trained and tested in completely different environments to show the capabilities of the model.

*Diyi Liu, Lanmin Liu, and Lee D Han*

1. **INTRODUCTION**

Ramp metering reduces overall freeway congestion by installing traffic signals on freeway on-ramps to manage the amount of traffic entering the freeway. Ramp metering strategy has been proven to be an effective method for decades to reduce traffic delays by decreasing speed variance, shockwaves, average delays, etc. The process of ramp metering is: (1) vehicles are fully stopped before stop bar; (2) traffic signal turns green; (3) once signal turns green, the drivers accelerate the vehicle to merge onto the mainline freeway. In **Figure 1**, the bottom right picture shows the on-ramp lane of a ramp metering site along the I-101 highway in California. In US, the ramp metering is applied by the "one car per greens" policy, which means that at most only one car can cross the stop bar every time there is the signal turns green. Therefore, since each red and green phase forms a signal cycle, the on-ramp traffic inputs are metered by varying the lengths of red cycle (thus the cycle length). Furthermore, detectors are needed to gather information to control the traffic. Until now, the most widely accessible detectors are to help ramp metering control are the loop detectors or its comparatives (e.g, radar detector). In this study, for each ramp metering site, four groups of detectors are as assumed: (1) upstream loop detectors; (2) on-ramp loop detectors; (3) on-ramp detectors, and (4) downstream detectors.

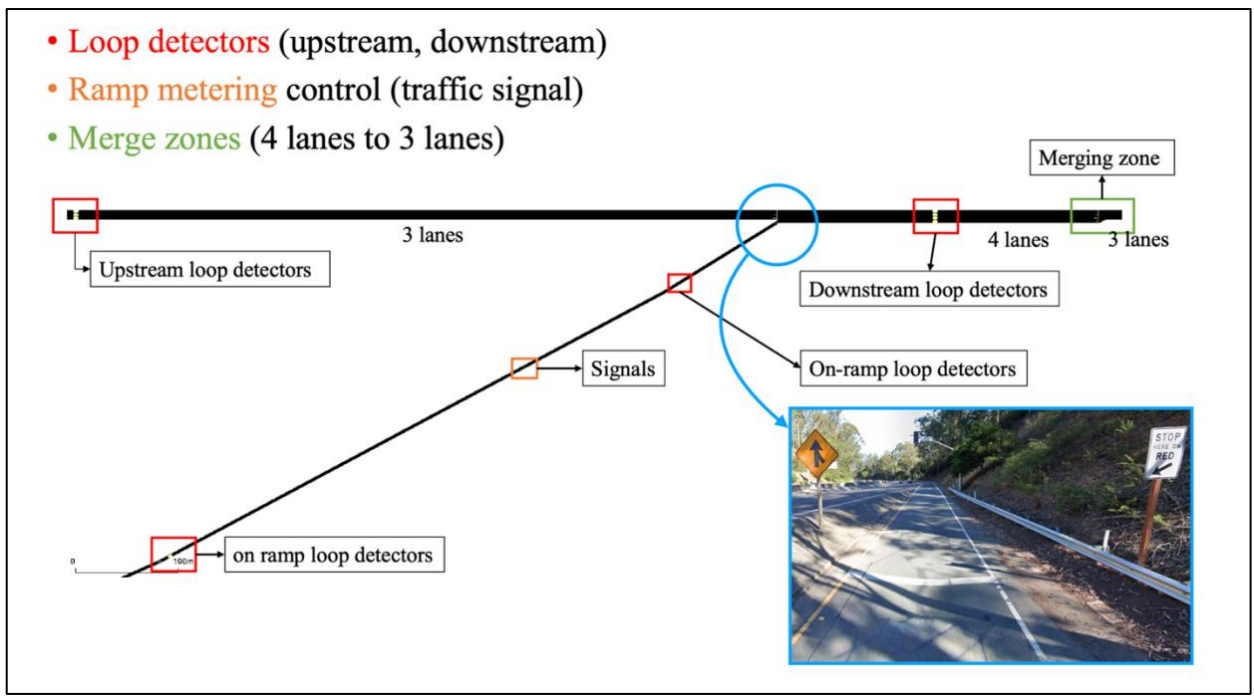

**Figure 1 Geometry Layout of a typical ramp metering site.**

Recently, reinforcement learning has become very successful in tackling many useful tasks including play games, robotics control, financial engineering, etc. Moreover, many applications like traffic signal control, connected/automated vehicles are all benefited from reinforcement learning. For reinforcement learning, in general, neither the programmer directly implements the strategy, nor the program/agent only learnt from labels. The core idea of reinforcement learning is to let the agent/program explore the possibilities by exploring different available actions over time to achieve some objectives with the hope that a strategy is found by trial and error through some feedback loops (by using the rewards function).

**Figure 2** describes the basic idea of reinforcement learning. **Figure 2** (a) splits a system into two parts: an agent and the environment. Given time $t$, the states $s_t$ are the numbers describing the state of the environment. The states $s_t$ are information collected by the available detectors. For any agent, it must interact with the environment it lives in for its benefits through some actions $a_t$. Finally, the rewards function $R(s, t)$ should be formulated the immediate benefits at the current state and time. The loop of sensing states and taking actions repeats over time until the experiment ends. Such an experiment formed a trajectory of actions and states, as demonstrated in **Figure 2** (b). Notice that the rewards function usually describes the direct benefits, and maximizing the long-term rewards are the objective for long term benefits. Thus, Q-value function is described as the exponential summation of rewards over the trajectory of the experiment.

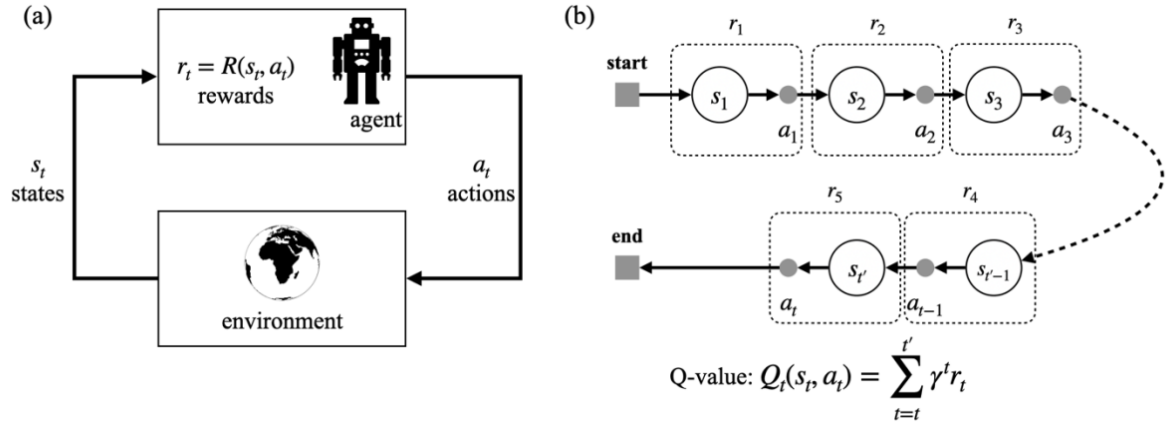

**Figure 2 The basic idea of reinforcement learning explained in the contexts of Markovian Decision Process.**

## 2. LITERATURE REVIEW

A considerable amount of literature has been published on ramp metering strategies and algorithms, which could generally be divided into three categories: fixed time, local control, and system-wide control. Papageorgiou [1] concludes that the strategies used for ramp metering: (1) fixed-time strategies; (2) Reactive strategies; (3) Nonlinear optimal Ramp metering strategies; (4) integrated freeway network traffic control. A freeway simulation was conducted in the study to compare densities/queues results between cases of control and no control.

Fixed time metering is the simplest approach with fixed cycle length, but it is also considered low efficient because the metering rate couldn't be adjusted according to the real-time freeway traffic states. System-wide control is proper when it comes to system optimization, which is responsive to corridor-wide real-time traffic conditions. System-wide control is usually based on local control except that multiple ramps along the corridor are considered at the same time. ALINEA is one of the local control strategies proposed by Papageorgiou [2] in 1991. In this paper, the metering rates are modeled as a control theory problem, which is determined based on occupancy data collected from mainline loop detectors located downstream to maximize the mainline throughput. An experimental study was implemented in Paris, France. Although the ALINEA method becomes the most recognized one, some limitations remain. First, the downstream bottleneck cannot be too far away from the ramp's site suffering from the "poorly damped closed-loop behavior". Second, the critical occupancy needs to be estimated. Third, the placement of the loop detector must be at the traffic bottlenecks. Since the introduction of ALINEA, many similar variants are proposed to improve its performance and applicability. For example, instead of measuring the downstream location, AU-ALINEA[3] used the measurements from the upstream site

*Diyi Liu, Lanmin Liu, and Lee D Han*

instead of the downstream. PI-ALINEA [4] is proposed to tackle different geometry cases with satisfactory performance including an uphill case, a lane drop case, and an "uncontrolled downstream on-ramp case". Notice that although ALINEA method becomes a successful one, there are a number of other traditional control methods proposed to solve ramp metering. For example, Gomesa [5] proposed the a cell transmission model (CTM) for optimizing ramp metering. A statistical model was applied by Ma [6] to evaluate the effectiveness of the before/after the ramp.

Recently, reinforcement learning has become another useful method to train algorithms in ramp metering due to the advancement in computing power. Rezaee [7] applies reinforcement learning uses the KNN-TD method to represent continuous state space. Three control methods are compared built test beds to test the performance [8]. Schmidt-Dumont [9] uses reinforcement learning (Q-learning) for optimal control, in which state-action q-values are predicted using a neural network instead of a table. Belletti [10] tested the reinforcement learning algorithm by simulation as well as generating a spatiotemporal diagram to show the speed distributions. From a system-wide aspect, Lu [11] considered minimizing TTT and penalty using the variation of variables for equity issues (e.g., queue length) to solve multiple ramp metering problems and did a simulation using a real network layout. Vrbanic [12] discussed VSL and RM together considering the involvement of autonomous vehicles, connected vehicles, etc.

While many reinforcement learning algorithms are claimed to be effective, the performance of these models cannot be guaranteed for several reasons: (1) some work trained and tested a network over the same geometry network with similar demands, which may lead to overfitting model; (2) as the states increased as a higher dimensional vector, it becomes easy to pollute the model either through False Data Injection Attacks or just through random noises; (3) there lacks a uniform formulation of states and values for all ramp metering geometries and lane layouts. In this study, a uniformed formulation that can be applied on all ramp metering sites are proposed. Moreover, different training and testing environments are used to show the effectiveness of the model. Finally, the model is attacked using a kind of False Data Injection attack, namely the Fast Gradient Sign Method (FGSM) attack.

## 3. METHODOLOGY
### 3.1 Formulating the ALINEA algorithm

ALINEA is a real-time ramp metering strategy that controls the ramp input traffic flow by monitoring the traffic occupancy on the mainstream at the downstream bottleneck. During the run, ALINEA keeps calculating the metering rate ($\tilde{r}$) over time with the objective of maximizing the throughput at the bottleneck. (The tilt hat in the symbol $\tilde{r}$ is to distinguish with the rewards function in reinforcement learning formulation in section 3.2). The normal scenario for ALINEA requires that a traffic signal is installed on the ramp which is to control the ramp input traffic and loop detectors that are installed downstream of the main road, which is to measure the occupancy of the mainstream. The formulas are shown below:

$$\tilde{r}(k) = \tilde{r}(k-1) + K_R * (\hat{o} - o_{ds}(k-1))$$
$$\tilde{r}(k) = \begin{cases} r_{min}, if\ \tilde{r}(k) > r_{max} \\ r_{max}, \tilde{r}(k) < r_{min} \end{cases}$$

where $k = 1, 2, 3,$ is the cycle index; $\tilde{r}(k)$ is the metering rate at the cycle $k$; $K_R$ is a fixed parameter controlling the sensitivity; $\hat{o}$ is the desired occupancy where it is assumed that traffic flow is maximized at this site; $o_{ds}$ is the downstream occupancy detected by loop detectors; $r_{min}$ and $r_{max}$ are the minimum and maximum ramp metering rate, respectively. In this study, the maximum and minimum metering rate ($r_{min}$ and $r_{max}$) correspond to 1600 vehicles/hour and 400 vehicles eh/hour, respectively.

*Diyi Liu, Lanmin Liu, and Lee D Han*

**3.2 Formulating the Deep Q-Learning algorithm**
*3.2.1 the overall deep q-learning framework*

A very general introduction of reinforcement learning is depicted in the Introduction section. In this section, reinforcement learning is formulated for the ramp metering control case. Formally speaking, reinforcement learning can be formulated using a Markov Decision Process (MDP) giving the following 5 tuples:

$$[T, A, S, r_t(s_t, a_t), P_t(s_t|s_t, a_t)]$$

where $T = \{1, \ldots, t'\}$ is the set of time steps during the control process; $A \in \{a_0, \ldots, a_{m-1}\}$ corresponds to the set of all available actions of the agent; $S \in \{\vec{s}|\vec{s} = \{d_1, \ldots, d_n\}\}$ is the set of all possible vector inputs $\vec{s}$, which is the results of $n$ signals generated by the detector's signal. $r_t(s_t, a_t)$ is the rewards function given time $t$, assuming the environment state $s_t$ and the taken action $a_t \in A$. Finally, $P_t(s_{t+1}|s_t, a_t)$ is the transition probability from time $t$ to $t+1$.

In this study, Q-Learning algorithm is employed, and the corresponding Bellman equation for updating Q-values are given as follows:

$$Q(s_t, a_t) = (1-\alpha)Q(s_t, a_t) + \alpha * (r_t + \gamma * max_a Q(s_{t+1}, a))$$

where $\alpha$ and $\gamma$ are two parameters controlling the behavior updating the q-value. $\gamma$ is the decay factor over the sequence of rewards $r_t$. $\alpha$ balances the weights between the last and this updating processes. In this study, $\alpha$ is set to 1 and $\gamma$ is set to 0.85. Since $s_t$ is a continuous space instead of discrete numbers, the states are used as input to a feed forward neural network, which generates the output of the fitted Q-values for all candidate actions. In the Bellman equation, the $max_a Q(s_{t+1}, a)$ just action corresponds to the "best" Q-values.

*3.2.2 formulating the rewards*

In this study, each agent corresponds to a ramp metering site (see **Figure 1**). In this study, the objective of the ramp metering control is to minimize the vehicle's Total Travel Time (TTT) during the simulation run. Also, it is assumed that the action (i.e., signal control plan) would work for a fixed time $\Delta t$(e.g., 30 seconds) before updating. Thus, the TTT between time t and t+1 becomes $\Delta t \times (N_{mainline}^i + N_{ramp}^i)$, and the rewards function can be simplified as:

$$r_t^i = -(N_{mainline}^i + N_{ramp}^i)$$

where $N_{mainline}^i$ and $N_{ramp}^i$ are just the number of observed vehicles on mainline and on the on-ramp for the $i$th ramp metering site, respectively. Simply speaking, the rewards are proportional to the negative of the total vehicles observed at the decision time. This formulation is also suitable as in practice the detector readings will only be updated every 30 seconds.

*3.2.3 formulating the states*

In this study, both the occupancy and the volume readings are used as the algorithm's input states. As mentioned, the detectors are grouped into four groups: (1) upstream loop detectors; (2) on-ramp loop detectors; (3) on-ramp detectors, and (4) downstream detectors. For each group, the total volume and the averaged occupancy are generated and aggregated every 30 seconds. For Deep Q-Learning, the aggregated records during the last 60 seconds are used. Thus, the input states are a vector of length $4 \times 2 \times 2 = 16$ states. This input state formulation can be adapted to different geometries. For example, some ramp metering site has two queuing/acceleration lanes. During the training process, we train the model on a corridor with different geometries. Thus, the model can be plugged-in directly for any ramp metering geometry.

### 3.3 Evaluating methods
There are three evaluating methods. First, to evaluate the model, the TTT is used as the first metric. Second, the speeds along the mainline corridor are also visualized to compare between different models. For the first two methods, three models are compared together: (1) no control; (2) ALINEA; and (3) Deep Q-Learning.

For the third method, the FGSM method is applied to attack/undermine the model. As a reference, the simplified steps of the FGSM attack are listed and explained as follows: (1) The attacker decides the target of interest (e.g., block a specific traffic lane or on-ramp); (2) Then, given output selected as the target of interest, a partial derivative is taken with respect to the input data to decide the gradient sign for each data input. A noise data is generated by taking a small step along each gradient sign direction. The FGSM method, by injecting a small value over the clean sample, generates the adversarial samples that trick the deep learning model to generate wrong outputs (the output desired by the attacker).

## 4. EXPERIMENTS
### 4.1 Training the Deep Q-Learning model
In this study, SUMO is used as the skeleton software to simulate the microscopic traffic behaviors. During the training, a straight corridor with 3 on-ramps are used to test the performance, as shown in **Figure 3**. The three ramps are all of different geometries. The first two ramps have two queueing lanes, whereas the third ramp only has a single queueing lane. The second ramp, unlike the other two, has two acceleration lanes, and the signal can release two vehicles at one green instead of only one vehicle. Finally, the first site suffers a weaving zone between the on-ramp and off-ramp, making it harder the mitigate conflits at that region.

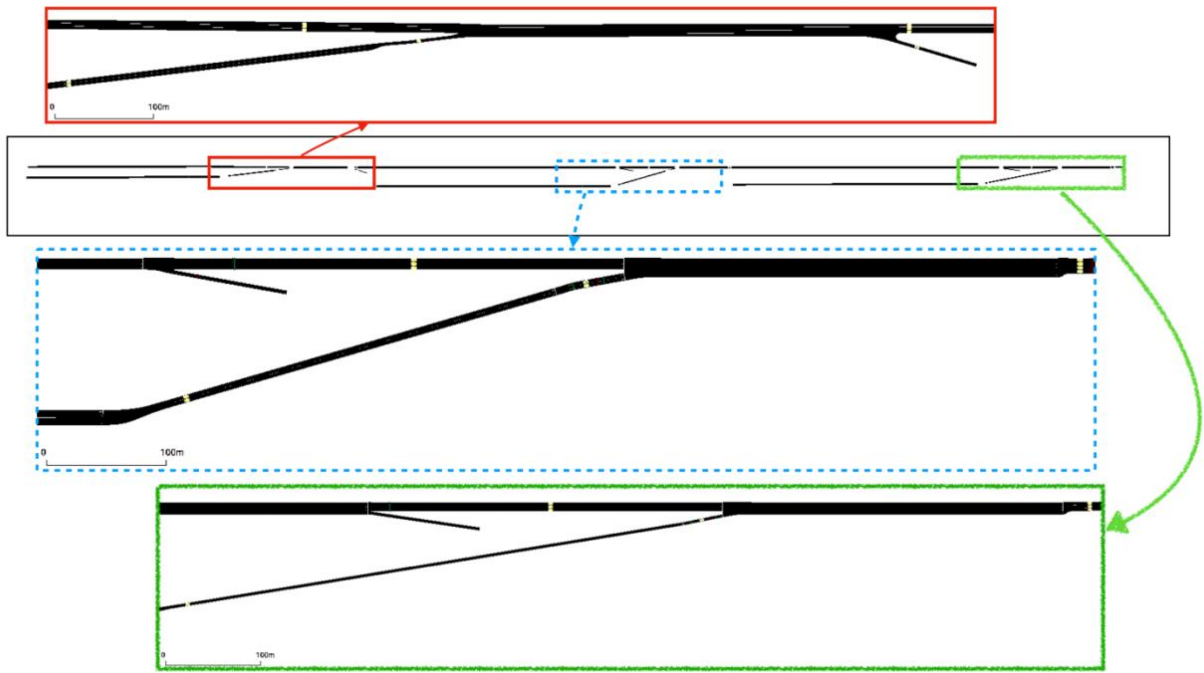

**Figure 3 The geometry layouts of the model training ground containing 3 on-ramps.**

*Diyi Liu, Lanmin Liu, and Lee D Han*

Besides configuring different geometry layouts, the traffic demand layouts are also randomized a lot to make the model more versatile and more robust towards fluctuations in traffic demands. For each training epoch every 15 minutes, a new traffic volume is sampled by picking one of the following 8 distributions in **Table 1**. Each simulation's time length is 2 hours and won't be terminated before all traffic dissipates.

| On-ramp's Demand ID (low to high) | Demands (veh/hour) | Mainline's Demand ID (low to high) | Distribution (veh/hour) |
|---|---|---|---|
| OR1 | $N(400, 20^2)$ | ML1 | $N(2,000, 40^2)$ |
| OR2 | $N(500, 25^2)$ | ML2 | $N(2,500, 50^2)$ |
| OR3 | $N(600, 30^2)$ | ML3 | $N(3,000, 60^2)$ |
| OR4 | $N(800, 40^2)$ | ML4 | $N(3,500, 70^2)$ |
| OR5 | $N(1000, 50^2)$ | ML5 | $N(4,000, 80^2)$ |
| OR6 | $N(1200, 60^2)$ | ML6 | $N(4,500, 90^2)$ |
| OR7 | $N(1300, 65^2)$ | ML7 | $N(5,000, 100^2)$ |
| OR8 | $N(1400, 70^2)$ | ML8 | $N(5,500, 110^2)$ |

**Table 1 Traffic demands to sample from during the training process.**

### 4.2 Testing and Comparing different models along the site

To test the model, different ramp metering models are applied a real corridor near Murfreesboro, TN. As shown in **Figure 4**, 5 consequtive on-ramps along I-24 near are selected to test the model's performance. The North/East-bound traveling direction (traveling inbound towards Nashville, TN) is tested.

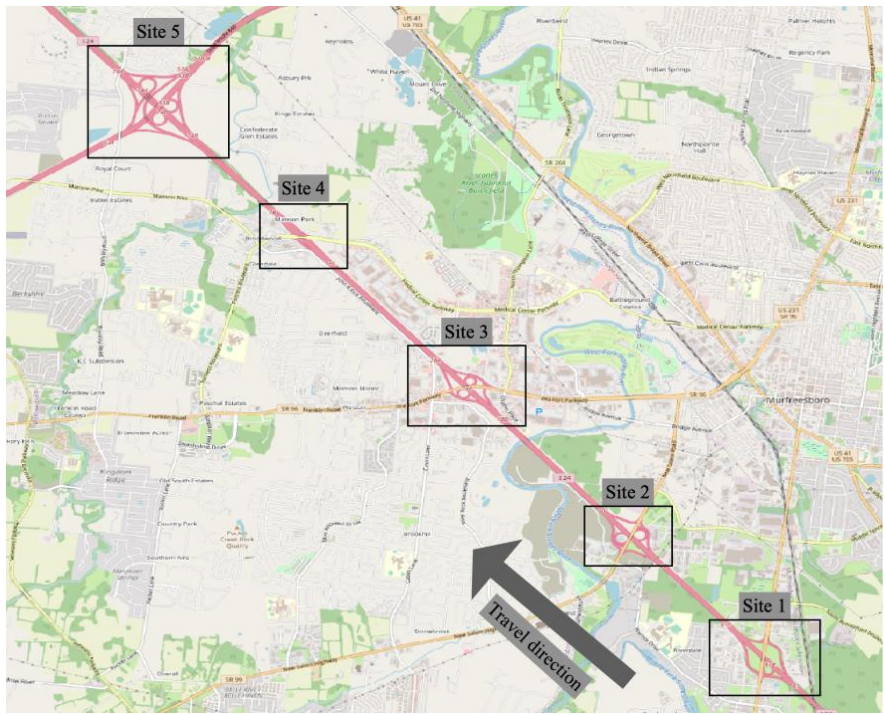

**Figure 4 The applied testbed for evaluating different ramp metering algorithms.**

Different from the training phase, a fixed testing demand is used to pressure test the system (see **Table 2** below). Notice that in **Table 2**, Site 1, 3, 4 used a fixed input demand whereas 2,5 used another set of demands. This is because site 2 and 5 has only one on-ramp queueing lane, whereas the other 3 on-ramp sites have two queueing lanes and two acceleration lanes.

| Time range | Demands (vehicle/hour) | | |
| --- | --- | --- | --- |
| (in minutes) | **Mainline** | **Site 1/3/4** | **Site 2/5** |
| 0-15 | 5,500 | 1,100 | 700 |
| 15-30 | 6,000 | 1,200 | 800 |
| 30-45 | 6,500 | 1,300 | 900 |
| 45-60 | 6,000 | 1,400 | 1,000 |
| 60-75 | 5,500 | 1,500 | 1,200 |
| 75-90 | 5,000 | 1,400 | 1,000 |
| 90-105 | 4,500 | 1,300 | 900 |
| 105-120 | 4,000 | 1,200 | 800 |

**Table 2 Demand used during the testing case.**

The most obvious way of comparing speed is by plotting them over space and time for different models. In **Figure 5**, the y-axis stands for the mileage over the traveling direction; x-axis stands for the simulation time between 0 and 120 minutes. For each 100 meters and 5 minutes, a rectangle is drawn with color showing the averaged traveling speed. By comparison, both the ALINEA and the Deep Q-Learning algorithm can enhance the traffic speed and mitigate congestions on the mainline. Compared with ALINEA method, it seems like Deep Q-Learning algorithm has significantly enhanced the traveling speed on the mainline. The last two columns of subfigures showed the cases where Deep Q-Learning algorithms are attacked using Fast Gradient Sign Method. On case named "Attack to Green" tries to block mainline by keeping the metering rate as high as possible. The other named "Attack to Red" tries to block to on-ramp by keeping the metering rate as low as possible. Both attack scenarios showed a high delay along the mainline.

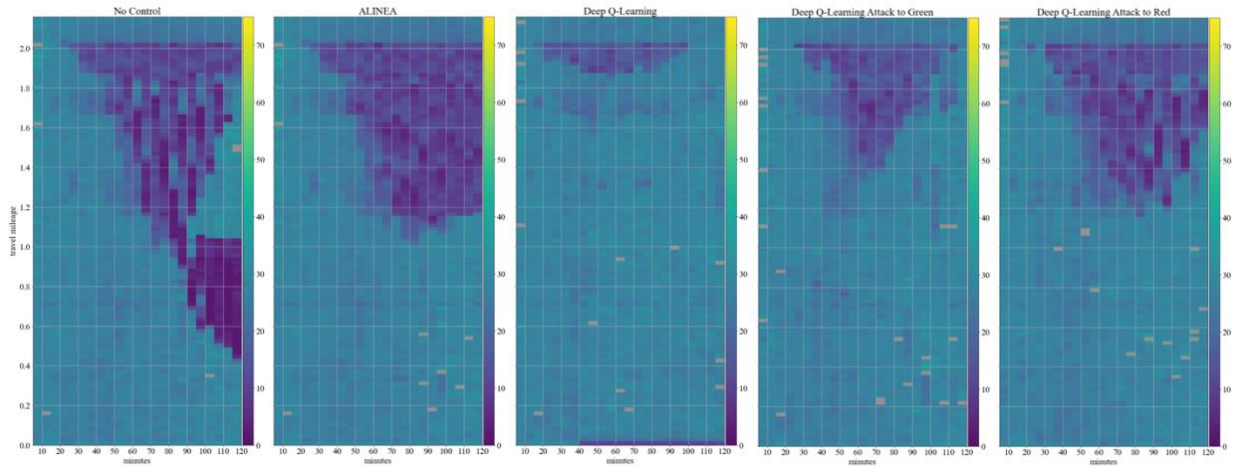

**Figure 5 A spatiotemporal speed plot comparing five different control and pollute algorithms.**

Note that besides mainline, traffic status over the ramp should also be considered. One can consider that by comparing TTT, not only on the mainline but the overall system. The results also showed that Deep Q-Learning method outperforms the other methods. Although the results get worse under two FGSM False Data Injection attacks, the overall TTT is still less than that of ALINEA.

|                      | TTT (in seconds) |           |           |           |           |           |
| Applied algorithm    | Overall    | Site 1    | Site 2    | Site 3    | Site 4    | Site 5    |
| No control           | 15,958,515 | 700,191   | 1,983,610 | 1,218,805 | 1,385,851 | 2,010,806 |
| ALINEA               | 14,880,797 | 702,608   | 761,873   | 916,411   | 1,774,126 | 2,047,778 |
| DQL                  | 12,594,484 | 1,186,593 | 1,030,121 | 995,028   | 983,724   | 1,735,887 |
| DQL (FGSM attack 1)  | 14,419,896 | 1,168,984 | 1,075,394 | 1,050,783 | 1,257,497 | 1,910,187 |
| DQL (FGSM attack 2)  | 14,602,978 | 1,160,074 | 1,019,047 | 771,400   | 1,510,706 | 2,097,751 |

**Table 3 Results of Total Travel Time**

## 5. CONCLUSIONS

In this study, we propose a Deep Q-Learning formulation for the ramp metering problem. Compared with previous studies, the method can be applied on different gemeotries and layouts by training with the model only once. The results of the proposed method outcompete the traditional ALINEA method as well as the No Control method. The Deep Q-Learning method is then attacked by FGSM applying a False Data Injection scheme. Even under attack, the results are still better than that of ALINEA.

However, there are still many things to be improved in this study. The authors acknowledge that there are a bunch of methods to control different sites corporately, especially the HERO algorithm. The team should also test the performance of those algorithms later. To talk about reinforcement learning, recently there are many more advanced methods come off the shelf. Those methods, typically the policy gradient methods, also shown a tendency of solving complicated problems, and one needs to formulate those methods and test them fors the application of ramp metering. Moreover, as for robustness analysis, there are other algorithms besides FGSM to be tested.

**ACKNOWLEDGMENTS**

The authors would like to thank for the National Science Fundation Project (Unique Entity Identifyer: FN2YCS2YAUW3), namely Secure Constrained Machine Learning for Critical Infrastructure CPS, for providing the opportunity and funding for investigating Machine Learning applications in transportation systems.

**AUTHOR CONTRIBUTIONS**

The authors confirm contribution to the paper as follows: study conception and design: Diyi Liu, Lanmin Liu; data collection: Diyi Liu, Lanmin Liu; analysis and interpretation of results: Diyi Liu, Lanmin Liu, Lee D Han; draft manuscript preparation: Lanmin Liu, Diyi Liu, Lee D Han. All authors reviewed the results and approved the final version of the manuscript.